\documentclass{article}
\usepackage[utf8]{inputenc}
\usepackage{preamble}

\title{Designing a Recurrent Neural Network to Learn a Motion Planner for High-Dimensional Inputs}
\author{Johnathan Chiu}
\affil{University of California, Berkeley}
\date{May 2022}

\begin{document}

\maketitle

\setlength{\abstitleskip}{-\absparindent}

\begin{abstract}
The use of machine learning in the self-driving industry has boosted a number of recent advancements. In particular, the usage of large deep learning models in the perception and prediction stack have proved quite successful, but there still lacks significant literature on the use of machine learning in the planning stack. The current state of the art in the planning stack often relies on fast constrained optimization or rule-based approaches. Both of these techniques fail to address a significant number of fundamental problems that would allow the vehicle to operate more similarly to that of human drivers. In this paper, we attempt to design a basic deep learning system to approach this problem. Furthermore, the main underlying goal of this paper is to demonstrate the potential uses of machine learning in the planning stack for autonomous vehicles (AV) and provide a baseline work for ongoing and future research.
\end{abstract}

\section{Introduction}

The goal in developing self-driving vehicles is to automate safe and efficient transportation for any passenger. When planning a safe route, a driver needs to consider a number of requirements before making any operation. To name a few of the more important requirements, a driver should avoid collisions, avoid driving recklessly, and follow traffic regulations. One of the bigger issues is, despite humans being able to learn to drive easily, not all humans drive in what we classify as "safe" mannerisms. Unsafe mannerisms is often denoted by actions of, including but not limited to, violating speed limits, making spontaneous lane changes in heavy traffic, and disregarding traffic lights and signs. Given these hazards in driving, motion planning systems in self-driving vehicles should be able to operate under all the aforementioned situations and more.

Traditional motion planners for autonomous vehicles are often encoded by a set rules hand-crafted by engineers for safe operation. These traditional methods require a significant number of edge cases to be accounted for and are not truly feasible in tackling the long-tail problem in the self-driving space. Likewise, other methods such as imitation learning have their own set of limitations that being the reliance on massive amounts of data to be generated by real drivers. This method can also often be costly since many companies are unwilling to share proprietary datasets. Finally, reinforcement learning methods require training on very specific maneuvers and relies on a master policy to make decisions on what maneuvers to take.

We turn to deep learning methods to attempt to tackle the problem. Specifically, we use a Recurrent Neural Network (RNN) to handle the controls over a series of states. We attempt to simplify the dimensionality of this problem by considering only important input parameters and relying on other parts of the AV stack that have proven to perform well. Specifically, rather than inputting the entire map, we only input information regarding road boundaries and the current lane we are in. Additionally, we encode only some number of the obstacles nearest to us rather than all of them in the scene. We can make these assumptions due to the abilities of current state of the art models in perception, prediction, and mapping. We further justify the safety and validity of this system despite removing numerous details in the scene in the following sections.

\section{Related Works}

A survey of the ongoing work from leading companies in the self-driving industry suggests that the motion planning problem in self-driving is far from being solved. Every researcher has their own separate approach to tackling the problem. In this section, we highlight a few related works from both industry and academia.

\paragraph{End-to-End Planning} Some of the previous literature suggests the use of an end-to-end method. In particular, researchers at NVIDIA designed a system that generates a set of controls directly from the provided images taken by onboard camera sensors \cite{end2end_2017}. Majority of the work nowadays have moved away from these ideas with emphasis placed on isolating different parts of the self-driving stack and tackling each component of the stack individually.

\paragraph{Reinforcement Learning} Reinforcement learning (RL) techniques often require an agent to explore an environment and develop optimal policies that enable it to navigate its surroundings. This requires the use of simulation and, during training, the data may not always reflect the distributions of real-life driving scenarios. Duan et. al \cite{hierarchical_rl_2020}, introduces a RL approach. But, again, this work relies on training very specific maneuvers which creates a complex system and may not generalize for edge cases that can appear more often than considered.

\paragraph{Imitation Learning} In Ashesh et al. \cite{woven_planet_2021}, the basis of the work relies on imitation learning -- using human demonstrations to train a model. This method is promising, but it requires a significant number of training samples from real-world data. This is costly and somewhat infeasible as it may limit the ability to scale for production. In addition, this methodology requires models with large numbers of parameters which could cause issues in situations where quick reaction times are required.

\paragraph{Constrained Optimization} Model Predictive Control (MPC) has been a longstanding solution to the motion planning problem. The ubiquity and power of current CPUs and GPUs have enabled this method to work well in some situations. While this is true, optimizing for a real-time solution could often fail when considering a significant number of input parameters found in real-world driving scenarios. 

Researchers at Baidu attempt to tackle this problem in Fan et al. \cite{apollo_baidu_2018}. The authors use an idea that combines dynamic programming (DP) and quadratic programming (QP). In another of their papers, they use a QP approach again to generate a system that is capable of avoiding collisions during motion planning \cite{quadratic_baidu_20202}.

\paragraph{Search Algorithms} Traditional search algorithms such as A$^*$ introduce the idea of using a discrete map and exploring all possible nodes before making a specific move. Though the exhaustive search is safest in theory, it is too slow for practical uses. On the other hand, Tesla's current path planning system uses Monte Carlo Tree Search (MCTS) which has proven to work well in a number of use cases \cite{tesla_fsd_2021}.

\newcommand{\matrixNot}[1]{\bm{\mathit{#1}}}

\section{Overview of System}

\subsection{Generalizing a Coordinate Frame}

We use a continuous 2-Dimensional Cartesian coordinate system to represent the position of the ego vehicle and the world around us. The ego vehicle is always considered to be the centerpoint in our coordinate system. This implies that all objects in the scene are always shifted about the ego vehicle. This coordinate frame allows the model to better learn and generalize since the initial position is fixed and are features that the model does not need to learn.

\subsection{Information from Higher-Level Systems}\label{sec:navsystem}

Our system relies on information provided from the perception module and a known high-definition road map of the world around us. This system also assumes that the vehicle's path planning navigation provides a route to take to navigate from point A to B with high detail. By high detail, we mean that the vehicle's navigation system provides which particular lane the vehicle should take on any given road. We further rely on the perception module to provide the polyline coordinates for the desired centerline (of the lane provided by the navigation) and road boundaries in question. Finally, we assume that all information of extraneous obstacles/objects in our local vicinity are provided to us.

\section{Framework}

\subsection{Model Input}

The model should understand how to react and create a navigable path given any current state of the vehicle. The state input to the model is a set of features containing its inital velocity, initial heading, and desired velocity. This is defined as a vector $ X_{input} = [v_0, h_0, v_d] $. As a reminder, the initial position, x, y-coordinates, can always be omitted since we shift the coordinate grid to center our vehicle. $v_0$ and $h_0$ are the initial velocity and initial heading, respectively. $v_d$ is the desired velocity which we assume to be determined by a blackbox system that determines the safest driving speed using the speed limit and road condition(s) at the time of operation.

We consider two additional pieces of information from the surrounding environment -- road lanes and objects/obstacles. We describe these in detail in the following sections.

\subsubsection{Polyline Inputs}

Our system only considers three pieces of road structure information: the centerline for the navigation system's desired lane to stay in, the road boundary on the left, and the road boundary on the right. We represent each piece of road/lane information as a polyline. Any polyline $\mathcal{P}$ is sequence of nodes. In our vector representation, $\mathcal{P}$ is simply $[p_1, p_2, ... , p_n]$ where $p_i$ is a x, y-coordinate pair $(x_i, y_i)$. The number of polyline coordinates, $n$, can be chosen dependent on how granular the input of the road should be.

\begin{figure}[hbt!]
    \centering
    \includegraphics[width=1.0\textwidth]{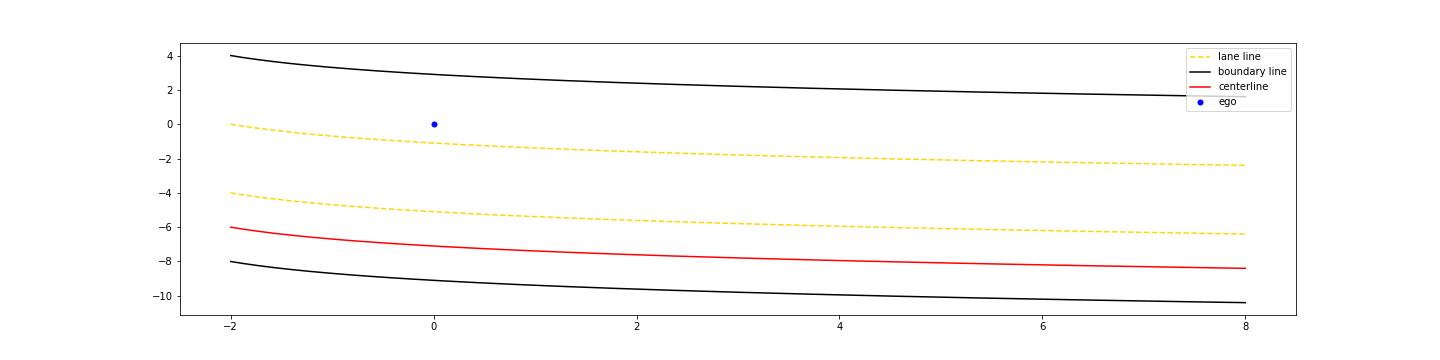}
    \includegraphics[width=1.0\textwidth]{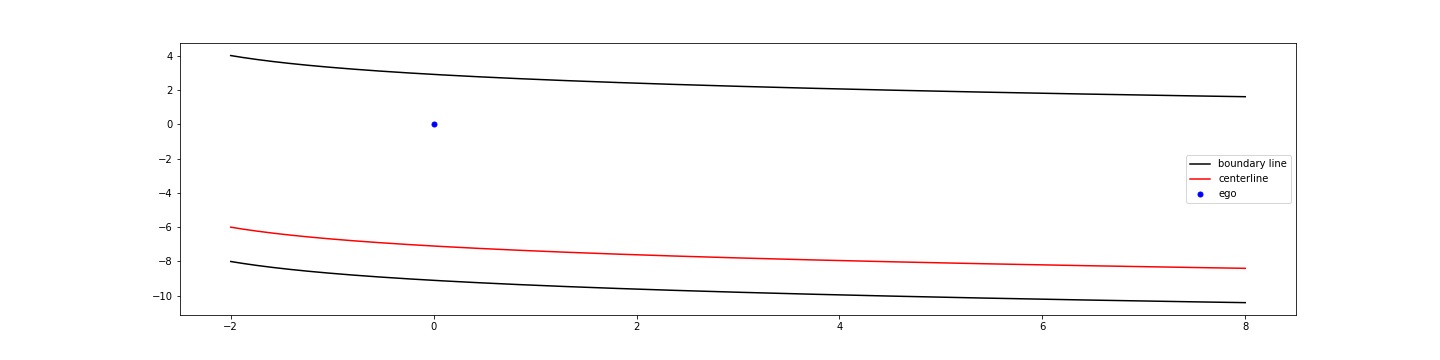}
    \caption{(\textit{Top}) Complete road information. (\textit{Bottom}) Model input.}
    \label{fig:laneinfo}
\end{figure}

As an example of our lane information, if we consider a 3-lane road and our navigation system suggests the vehicle to take the rightmost lane, the provided information is the left and right road boundaries and the centerline for the rightmost lane. The left lane and center lane information are not included in the system. Like mentioned in section \ref{sec:navsystem}, we rely on the navigation system to provide the motion planner with a specific centerline to follow. We designed our motion planner in this specific way to minimize the amount of information we need to consider and still operate under safety constraints. We further elaborate on this design methodology in section \ref{sec:decisions} and this specific example is illustrated in Figure \ref{fig:laneinfo}.

\subsubsection{Object Representation}

We represent the objects around us by encoding the positions of the $k$-nearest objects, $\mathcal{O}$, around us. We choose $k$ arbitrarily for testing but, in reality, this value should be large enough to encapsulate the number of obstacles in any true driving scenario. The objects surrounding the ego vehicle are represented by a vector of x, y-coordinates pairs identical to the polyline representation described above. In sparse driving conditions, such as at night, the vehicle may not encounter $k$ objects and does not need a vector of size $k$ for the surrounding objects/obstacles. In such situations, all leftover slots in the vector are filled with a $(0, 0)$ pair.

With this representation of objects, we can only input static features. We suggest that all objects be encoded with motion vectors dependent on the output from the motion prediction stack. We consider dynamic obstacles to be out of scope for this project and experiment with only static objects.

\subsection{Model Architecture}

Our system comprises of three separate networks. The first two of the networks are multi-layer perceptrons (MLP) used for embedding the road information and the object instances. The third network is a RNN which incorporates the embedded polylines (road information) and obstacles to plan a path for the vehicle. The planner should compute a series of controls -- acceleration and steering -- for some number of steps into the future, $\mathcal{T}$. Our output is represented as a sequence of controls: $Y_{control} = [(a_1, \dot\theta_1), (a_2, \dot\theta_2), ... , (a_T, \dot\theta_T)]$. We also have the model output values between [-1, 1] by applying Tanh activation function after the final layer. This allows us to explicitly define the acceleration and steering constraints outside of the model by multiplying the control value by its constraint value. For e.g., if the vehicle has an acceleration constraint of $3 m/s^2$ and a turning constraint of $40 ^{\circ}$, we multiply each $a_i$ by 3 and each $\dot\theta_i$ by 40 to impose the constraints.

\subsection{Loss Function}\label{sec:objfn}

We define our loss to consider a few factors to optimize over. The motion planner should avoid all objects surrounding the vehicle, drive at the speed of the desired velocity, avoid the road boundaries, and follow the navigation system's centerline (the vehicle's guiding path to get to a final destination). We incorporate the direction/heading of the centerline to ensure that the vehicle does not drive in the wrong direction. Additionally, the loss term at time $t$ is scaled by $t$. We do this to allow the model to slowly converge to a specific solution and allow it to implicitly explore "backtracking" options. Essentially, our objective function boils down to:

\begin{equation}
  \begin{aligned}[b]
      \mathcal{L}_{planner} =
      \sum_{t=1}^{\mathcal{T}} t (\mathcal{L}_{maneuver}(t) + \mathcal{L}_{safety}(t))
  \end{aligned}
\end{equation}

\noindent where 

\begin{equation}
  \begin{aligned}[b]
      \mathcal{L}_{maneuver}(t) = \alpha \mathcal{E}_{cte}(t) + \beta \mathcal{E}_{he}(t) +
      \gamma \mathcal{E}_{ve}(t)
  \end{aligned}
\end{equation}

\noindent and

\begin{equation}
  \begin{aligned}[b]
       \mathcal{L}_{safety} =  \mu \mathcal{E}_{collision}(t) + \rho\mathcal{E}_{boundary}(t) .
  \end{aligned}
\end{equation} 
\noindent $\mathcal{E}_{cte}$ is a function that describes the vehicles shortest distance from the centerline at time $t$, $\mathcal{E}_{he}$ is a function that returns the heading error between the vehicle's heading and the heading of the road at time $t$, and $\mathcal{E}_{ve}$ is the velocity error at time $t$. Additionally, $\mathcal{E}_{collision}$ returns the summed distance from the ego to all $k$-nearest neighbors at time $t$. The formula describing $\mathcal{E}_{collision}$ is

\begin{equation}
  \begin{aligned}[b]
    \mathcal{E}_{collision}(t) = \sum_{o \in \mathcal{O}} e^{5 - o_d}
  \end{aligned}
\end{equation}

\noindent where $o_d$ is the distance between the ego and object $o$ at time $t$. We use a shifted and scaled exponential function to imply that a small distance to objects should be classified as "reckless" driving and result in a larger loss. The shifting and scaling coefficients were determined empirically. Another thing to address is that we do not consider the distance of any objects with position $(0, 0)$ (the same as the ego) since we mentioned above these are just placeholders in the vector and not truly objects.

In addition, $\mathcal{E}_{cte}$ is defined as such:

\begin{equation}
  \begin{aligned}[b]
    \mathcal{E}_{cte}(t) = \min_{p_i, p_{i+1} \in \mathcal{P}} dist((x_t, y_t), (midpoint(p_i, p_{i+1}))
  \end{aligned}
\end{equation}

\noindent where $dist$ and $midpoint$ are the formulas to compute the Euclidean distance between two points and finding the midpoint between two points, respectively. In our implementation, we compute this by looping through each consecutive pair of polyline points while calculating the distance of the vehicle from the segment's midpoint and returning the smallest of distances between the vehicle position. Similarly, $\mathcal{E}_{he}$ uses the two points, $p_i$ and $p_{i+1}$ determined by $\mathcal{E}_{cte}$ to compute the relative heading of the road.

\section{Training}

During training, the models are trained in unsupervised manner. We use Adam optimizer to train the model's weights with a fixed learning rate of .00001. Additionally, we set the parameters of the loss function as follows: $\alpha = 15$, $\beta = 9$, $\gamma = 11$, $\mu = 100$, $\rho = 2\alpha$. Lastly, we train the model over 500 epochs.

Using unsupervised methods allow us to generate massive amounts of input data easily. We randomly synthesize a large number of initial states, road maps, and objects to train the model. We observed that the model was able to determine a satisfactory path using the provided objective function described in section \ref{sec:objfn}. We also suggest that in practice, the model should be trained with some number of supervised samples. We believe that using semi-supervised paradigms will allow the model to converge faster as it aids in finding satisfactory local minimums. Additionally, semi-supervised learning enables the model to learn directly from examples of human driving and attempt to mimic the actions and "thought processes" behind deciding which path to make while driving.

To better understand the model’s decisions during training, we also used the model solely as an optimizer over a single set of states. We found that the model tended to converge quickly, within 300 to 400 iterations, even in complex situations. Additionally, we observed that the model made interesting decisions in unusual scenarios. For instance, when we pointed the vehicle’s heading almost orthogonal to the road’s heading and near the boundary line, it learned to first reverse towards the centerline to adjust its heading before proceeding to move forward. We found this observation most interesting given that our loss function accounts for velocity error. The model was able to foresee that the loss becomes minimized with this action first. We believe this is as a result of scaling each term at time $t$ by the value of $t$.

\section{Design Decision Making}\label{sec:decisions}

Now that we have introduced our loss function in \ref{sec:objfn}, we continue our discussion of incorporating only a single centerline and disregarding other lane information. The main part of this design methodology was to simplify the inputs to the model while enabling us to completely remove rule-based operations for the vehicle. The motion planner should be able to independently solve all scenarios given enough training data. One example would be passing a slow vehicle ahead. In many current motion planners, there is a large reliance on a specific rule that determines when to pass a vehicle by moving into a different lane. In our design, the vehicle will automatically trigger a safe lane change if there is enough space on the left or right of the vehicle since the desired velocity is not achieved. The vehicle passes around the slow vehicle, being sure to avoid other impediments, and returns back to its original lane to minimize the distance to the centerline which is described in the loss function. 

Another example would be automatically changing lanes from the rightmost lane to left lane in preparation for a left turn. In this situation, the navigation system will relay information of a new centerline to the model, the left lane's centerline. Rather than explicitly telling the vehicle to make two consecutive lane changes from a rule-based planner, our system should determine when and how to make a safe lane changes to minimize the vehicle's distance to the provided centerline. Making the left turn works similarly -- we map a centerline of the turn lane and the the vehicle should make a turn that minimizes the distance between itself and the centerline of the turn lane.

A third interesting example that justifies our design would be how the vehicle should react at a red light. In rule-based motion planners, the vehicle is told to have a hard stop at a certain point. In our design, we set the desired velocity to 0 m/s and a red light can be denoted by a row of static "obstacles" before the intersection. The model will slow down and not cross the "obstacles" since there is no path around them the vehicle can safely take. This is another example of why our model can be more efficient than rule-based decision where engineers need not handcraft every rule for all edge cases.

\section{Results}

We perform a small case study by simulating a few complex scenarios (including some of the aforementioned scenarios in the previous section) and our model's outputted trajectories in response to the environment it is placed in.

\begin{figure}[hbt!]
    \centering
    \begin{subfigure}[t]{.475\textwidth}
        \centering
        \includegraphics[width=\linewidth]{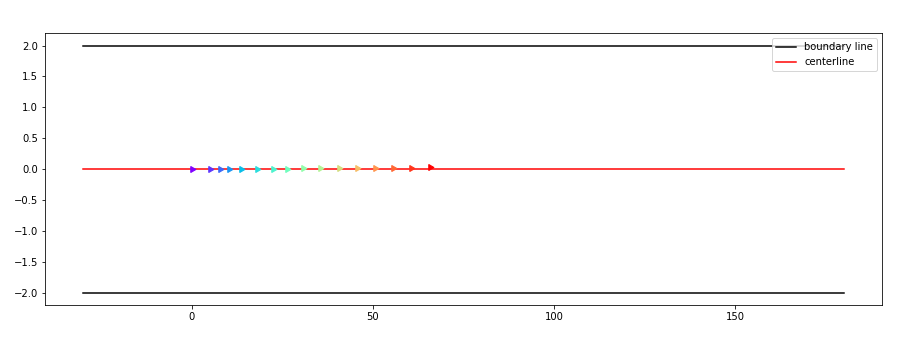}
        \caption{Ability to follow centerline. \textit{Initial Velocity}: $5 m/s$, \textit{Initial Heading}: $0^{\circ}$, \textit{Desired Velocity}: $5 m/s$}
        \label{fig:follow-lane}
    \end{subfigure}
    \hfill                      
    \begin{subfigure}[t]{.475\textwidth}
        \centering
        \includegraphics[width=\linewidth]{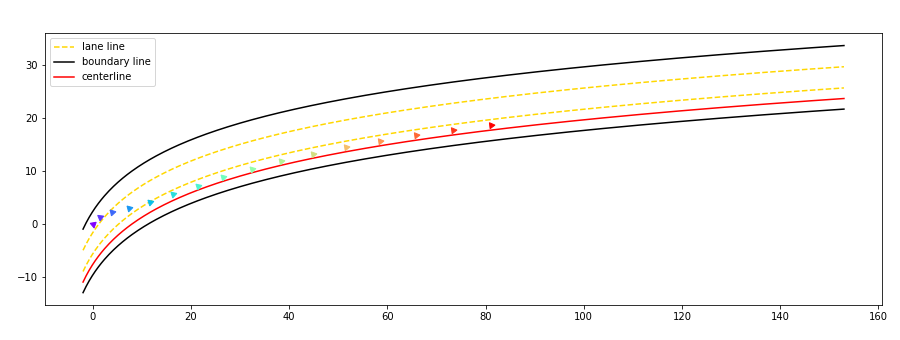}
        \caption{Changing lanes from left to right. \textit{Initial Velocity}: $3 m/s$, \textit{Initial Heading}: $40^{\circ}$, \textit{Desired Velocity}: $8 m/s$}
        \label{fig:change-polyline}
    \end{subfigure}
    \begin{subfigure}[t]{.475\textwidth}
        \centering
        \includegraphics[width=\linewidth]{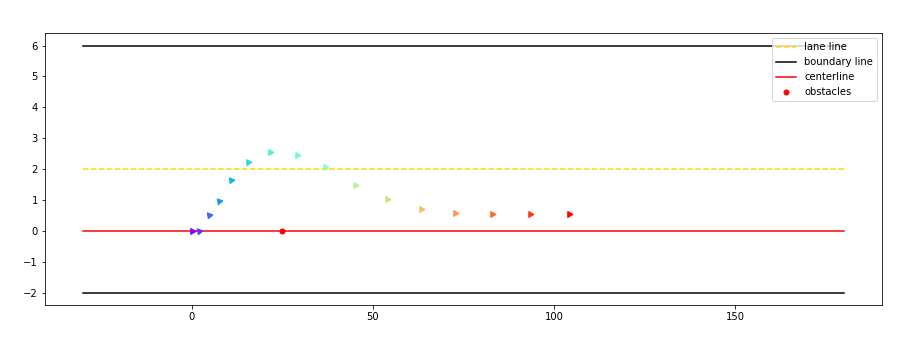}
        \caption{Moving around object directly ahead. \textit{Initial Velocity}: $2 m/s$, \textit{Initial Heading}: $0^{\circ}$, \textit{Desired Velocity}: $7 m/s$}
        \label{fig:go-around}
    \end{subfigure}
    \hfill
    \begin{subfigure}[t]{.475\textwidth}
        \centering
        \includegraphics[width=\linewidth]{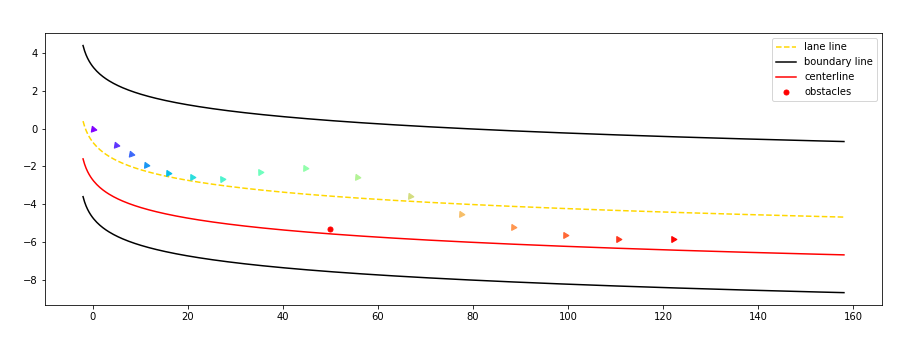}
        \caption{Changing lanes while avoiding object. \textit{Initial Velocity}: $5 m/s$, \textit{Initial Heading}: $0^{\circ}$, \textit{Desired Velocity}: $10 m/s$}
        \label{fig:go-around2}
    \end{subfigure}
    \begin{subfigure}[t]{.475\textwidth}
        \centering
        \includegraphics[width=\linewidth]{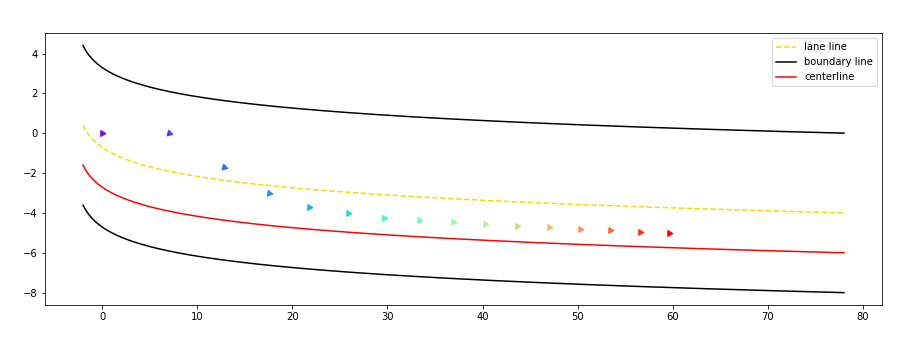}
        \caption{Another lane change example. \textit{Initial Velocity}: $10 m/s$, \textit{Initial Heading}: $0^{\circ}$, \textit{Desired Velocity}: $3 m/s$}
        \label{fig:lane-change}
    \end{subfigure}
    \hfill
    \begin{subfigure}[t]{.475\textwidth}
        \centering
        \includegraphics[width=\linewidth]{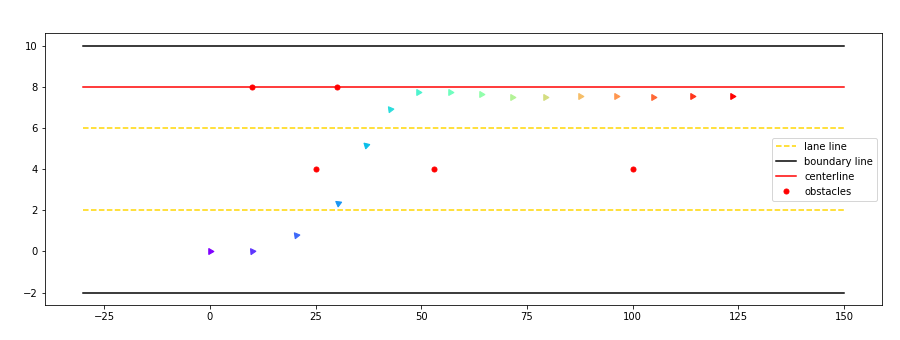}
        \caption{Avoiding objects while moving from right to left. \textit{Initial Velocity}: $10 m/s$, \textit{Initial Heading}: $0^{\circ}$, \textit{Desired Velocity}: $10 m/s$}
        \label{fig:left-to-right}
    \end{subfigure}
    \begin{subfigure}[t]{.475\textwidth}
        \centering
        \includegraphics[width=\linewidth]{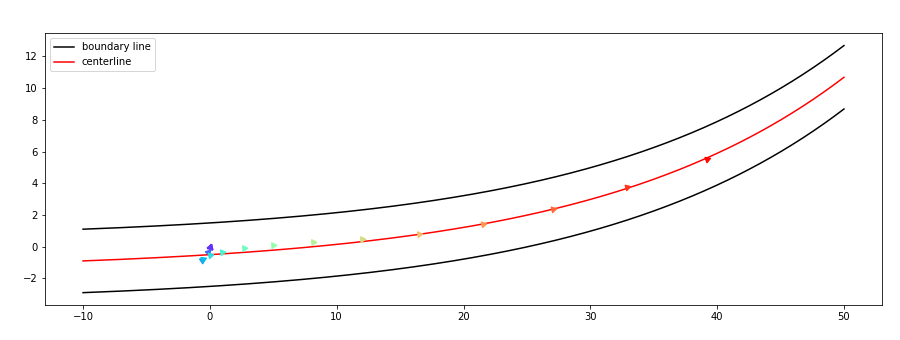}
        \caption{Reversing to correct heading before proceeding. \textit{Initial Velocity}: $0 m/s$, \textit{Initial Heading}: $0^{\circ}$, \textit{Desired Velocity}: $7 m/s$}
        \label{fig:reverse-forward}
    \end{subfigure}
    \hfill
    \begin{subfigure}[t]{.475\textwidth}
        \centering
        \includegraphics[width=\linewidth]{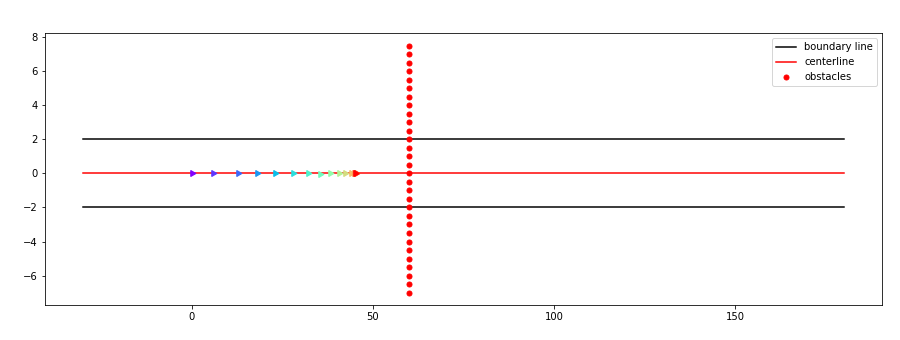}
        \caption{Stopping before a "red light". \textit{Initial Velocity}: $6 m/s$, \textit{Initial Heading}: $0^{\circ}$, \textit{Desired Velocity}: $0 m/s$}
        \label{fig:stop}
    \end{subfigure}
\end{figure}

\section{Conclusion}

We introduce a deep learning architecture and loss function for self-driving motion planners. The philosophy behind our design is to model the path planner to make decisions in similar ways that a human driver would. Our model demonstrates strong preliminary results due to its ability to handle a variety of driving situations/environments. Additionally, our model is able to learn to maneuver in situations that are often dictated by rule-based algorithms suggesting that rule-based models can be eliminated fully.

Some next steps in further developing our work would be to combine our model with other methods such as MCTS. This idea uses our deep learning model to guide an initial path and capitalize on the backtracking ideas in MCTS to explore the search space. Likewise, the initial trajectories outputted from our model can be fed into other refining algorithms like constrained optimizers.

Finally, in the previous sections we hinted at using motion prediction vectors to introduce dynamic scenarios for our model. This more accurately encapsulates real-world driving. We also observed in Gao et. al  \cite{vectornet_2020} the use of a graph neural network, VectorNet, in motion prediction. We suggest concatenating our model with VectorNet to both predict and plan accordingly to an environment with moving vehicles, pedestrians, etc.

\printbibliography

@misc{tesla_fsd_2021, 
    title={Deep understanding tesla FSD part 3: Planning \& Control},     
    url={https://saneryee-studio.medium.com/deep-understanding-tesla-fsd-part-3-planning-control-9a25cc6d04f0}, 
    journal={Medium}, 
    publisher={Medium}, 
    author={Zhang, Jason}, 
    year={2021},
}

@misc{apollo_baidu_2018,
  doi = {10.48550/ARXIV.1807.08048},
  url = {https://arxiv.org/abs/1807.08048},
  author = {Fan, Haoyang and Zhu, Fan and Liu, Changchun and Zhang, Liangliang and Zhuang, Li and Li, Dong and Zhu, Weicheng and Hu, Jiangtao and Li, Hongye and Kong, Qi},
  title = {Baidu Apollo EM Motion Planner},
  publisher = {arXiv},
  year = {2018},
  copyright = {arXiv.org perpetual, non-exclusive license}
}

@misc{vectornet_2020,
  doi = {10.48550/ARXIV.2005.04259},
  url = {https://arxiv.org/abs/2005.04259},
  author = {Gao, Jiyang and Sun, Chen and Zhao, Hang and Shen, Yi and Anguelov, Dragomir and Li, Congcong and Schmid, Cordelia},
  title = {VectorNet: Encoding HD Maps and Agent Dynamics from Vectorized Representation},
  publisher = {arXiv},
  year = {2020},
  copyright = {arXiv.org perpetual, non-exclusive license}
}

@misc{woven_planet_2021,
  doi = {10.48550/ARXIV.2107.08142},
  url = {https://arxiv.org/abs/2107.08142},
  author = {Jain, Ashesh and Del Pero, Luca and Grimmett, Hugo and Ondruska, Peter},
  title = {Autonomy 2.0: Why is self-driving always 5 years away?},
  publisher = {arXiv},
  year = {2021},
  copyright = {Creative Commons Attribution 4.0 International}
}

@article{hierarchical_rl_2020,
	doi = {10.1049/iet-its.2019.0317},
	url = {https://doi.org/10.1049%2Fiet-its.2019.0317},
	year = 2020,
	publisher = {Institution of Engineering and Technology ({IET})},
	volume = {14},
	number = {5},
	pages = {297--305},
	author = {Jingliang Duan and Shengbo Eben Li and Yang Guan and Qi Sun and Bo Cheng},
	title = {Hierarchical reinforcement learning for self-driving decision-making without reliance on labelled driving data},
	journal = {{IET} Intelligent Transport Systems}
}

@misc{end2end_2017,
  doi = {10.48550/ARXIV.1704.07911},
  url = {https://arxiv.org/abs/1704.07911},
  author = {Bojarski, Mariusz and Yeres, Philip and Choromanska, Anna and Choromanski, Krzysztof and Firner, Bernhard and Jackel, Lawrence and Muller, Urs},
  title = {Explaining How a Deep Neural Network Trained with End-to-End Learning Steers a Car},
  publisher = {arXiv},
  year = {2017},
  copyright = {arXiv.org perpetual, non-exclusive license}
}

@INPROCEEDINGS{quadratic_baidu_20202,
  author={Zhang, Yajia and Sun, Hongyi and Zhou, Jinyun and Pan, Jiacheng and Hu, Jiangtao and Miao, Jinghao},
  booktitle={2020 IEEE Intelligent Vehicles Symposium (IV)}, 
  title={Optimal Vehicle Path Planning Using Quadratic Optimization for Baidu Apollo Open Platform}, 
  year={2020},
  volume={},
  number={},
  pages={978-984},
  doi={10.1109/IV47402.2020.9304787}}

\end{document}